\documentclass[runningheads]{llncs}
\usepackage[T1]{fontenc}
\usepackage{graphicx}
\usepackage{subcaption}

\usepackage{hyperref}
\usepackage{color}
\usepackage{amsmath}

\begin{document}
\title{Rare anomalies require large datasets: About proving the existence of anomalies}
%
\titlerunning{About proving the existence of anomalies}
%
\author{Simon Klüttermann\inst{1}\orcidID{0000-0001-9698-4339} \and
Emmanuel Müller\inst{1}\orcidID{0000-0002-5409-6875}}

\authorrunning{S. Klüttermann et al.}
%
\institute{TU Dortmund University, Dortmund, Germany \\
\email{simon.kluettermann@cs.tu-dortmund.de}}

\maketitle              
\begin{abstract}
Detecting whether any anomalies exist within a dataset is crucial for effective anomaly detection, yet it remains surprisingly underexplored in anomaly detection literature. This paper presents a comprehensive study that addresses the fundamental question: \textit{When can we conclusively determine that anomalies are present?} Through extensive experimentation involving over three million statistical tests across various anomaly detection tasks and algorithms, we identify a relationship between the dataset size, contamination rate, and an algorithm-dependent constant \( \alpha_{\text{algo}} \). Our results demonstrate that, for an unlabeled dataset of size \( N \) and contamination rate \( \nu \), the condition \( N \ge \frac{\alpha_{\text{algo}}}{\nu^2} \) represents a lower bound on the number of samples required to confirm anomaly existence. This threshold implies a limit to how rare anomalies can be before proving their existence becomes infeasible. 

\end{abstract} 
\section{Introduction}
Anomaly detection is a diverse field encompassing numerous algorithms, ranging from relatively simple statistical approaches~\cite{lof,ifor} to more sophisticated pattern extraction methods~\cite{deepsvdd,aean}. These algorithms are applied across a wide array of domains, addressing challenges in areas such as fraud detection~\cite{auto_appl_electionfraud,fraudapl,creditcardfraud}, fault detection~\cite{faultapl,faultapl2,machinefault}, and scientific research~\cite{auto_appl_particlephysics,mikuni,auto_appl_virusmutations}. Despite the diversity of these algorithms and their applications, a common characteristic among them is their output of anomaly scores for individual samples. In general, a higher anomaly score indicates a greater likelihood that a sample is anomalous, allowing for a ranking of dataset samples based on their anomaly levels. However, in practical applications, the ranking of anomalies may not be the primary concern. While identifying where an error has occurred is often important, determining \textbf{that} an error has occurred is a more fundamental question. In this paper, we demonstrate that addressing this question is far from trivial and, in many cases, more complex than simply ranking anomalous samples.

Anomaly detection methodologies can be broadly categorized by data availability: with labeled data (supervised~\cite{auto_supervised}), which makes detecting that anomalies are present trivial, or without any access to labeled data (unsupervised~\cite{surveyruff}). In the unsupervised case, identifying the presence of anomalies can be challenging, as the focus is often on rare sample identification. And rare events, represented in the tail of a distribution, may not always be anomalous. Thus in this work, we focus on the middle ground between both, where some normal data is labeled, but all anomalies remain unlabeled (weakly supervised or one-class classification~\cite{surveyOneClass}). This scenario is frequently encountered, as it avoids the need for extensive labeling of rare anomalies while providing a more precise search goal than purely unsupervised detection. 

Our approach leverages the presence of a labeled set of normal samples alongside an unlabeled dataset, allowing us to examine whether anomalies can be detected by assessing differences between these datasets. To explore this question, we construct a comprehensive set of anomaly detection problems, covering four orders of magnitude in dataset sizes and five orders of magnitude in contamination rates within the unlabeled test set. Our analysis employs four different statistical tests and five distinct anomaly detection algorithms.

Through this analysis, we identify a decision boundary that differentiates between scenarios in which anomalies can be reliably detected and those where detection is unfeasible. Our findings indicate that this decision boundary is influenced by both the contamination rate and the dataset size, and remains similar across all tested statistical methods and anomaly detection algorithms. Finally, we introduce a thought experiment to show that we see a similar behavior even when drastically simplifying our setup and removing both the anomaly detection algorithms as well as the statistical tests.

Our code is available at \href{https://anonymous.4open.science/r/nu-B62E/README.md}{anonymous.4open.science/r/nu-B62E}.


\section{Related Work}
Numerous algorithms have been proposed for weakly supervised anomaly detection~\cite{surveyruff,surveyzhao,surveyOneClass}. However, in this study, we examine both exceptionally many and large datasets, necessitating the use of highly efficient algorithms. To address this, we selected five algorithms that represent distinct detection principles. Isolation Forest~\cite{ifor} identifies anomalies by isolating them through an ensemble of decision trees; COPOD~\cite{copod} estimates copulas to describe the probability density; A Gaussian Mixture Model (GMM)~\cite{gmm} assumes that the data distribution can be approximated as a sum of multiple Gaussian components and searches for regions with low density. HBOS~\cite{hbos} takes a different approach by assuming feature independence and computing histograms within each feature, while PCA~\cite{pca} models normal samples as following a lower-dimensional representation.

Each of these algorithms computes an anomaly score, ranking samples by their level of anomalousness. However, an anomaly score of, for instance, $3.2$ does not directly indicate whether a sample is genuinely anomalous or not, as it lacks an absolute meaning. Typically, a threshold is applied, such as classifying samples with scores higher than $95\%$ of normal training samples as anomalous. Yet, in practice, the primary question may not be where an anomaly has occurred but whether any anomaly is present at all. For instance, in aviation, the occurrence time of an electronic fault in the last flight may be less crucial than knowing if a fault occurred at all. This binary determination cannot be achieved solely through a threshold-based approach, which usually designates a fraction of samples as anomalous regardless of the actual presence of anomalies.

To address this, research has explored methods for translating anomaly scores into probabilities~\cite{probaad}. However, this remains a challenging problem~\cite{compareOutlierProbabilities} as accurate probability estimation requires careful calibration, and determining whether anomalies are present would necessitate aggregating potentially thousands of probability values, thus compounding any errors. In contrast, we propose a different approach by directly testing for statistical differences between normal and potentially contaminated samples.

Our analysis uncovers a relationship between the minimum sample size required to detect anomalies and the rarity of the anomalies in question, which we formulate as \( N \ge \frac{\alpha}{\nu^2}, \; \alpha > 1 \). Interestingly, a recent study~\cite{magna} also noticed that the novel anomaly detection they introduced only works when a similar condition is fulfilled (\( N \gg \frac{1}{\nu^2} \)), suggesting that this relationship may be fundamental and applicable across various methods.

\subsection{Statistical Tests}
To determine whether a potentially contaminated dataset contains anomalies, we employ statistical tests. Specifically, we focus on tests that assess differences between a known uncontaminated reference dataset and a potentially contaminated, unlabeled test set. 
The null hypothesis assumes that the distributions of the two datasets are identical. Our objective is to identify cases where a statistically significant difference ($p<0.05$) indicates the presence of anomalies.  
Several statistical tests have been developed for this purpose, each with distinct advantages and limitations. In this work, we employ four such tests. First, we use the Kolmogorov-Smirnov test~\cite{kolgomorovsmirnov}, which compares empirical distribution functions. Next, we apply the Anderson-Darling test~\cite{andersondarling}, which is more sensitive to the tails of distributions and may be particularly effective for detecting outliers. Similarly, we consider the Mann-Whitney U test~\cite{mannwhitney}, known for its robustness to outliers, making it less prone to being misled by extreme values in the training set. Finally, we also include the simpler Student’s t-test~\cite{studentsTtest}.

\section{Methodology}

To analyze the effect of dataset size across multiple orders of magnitude, we rely on simulated data, as most real-world anomaly detection benchmark datasets contain only a few thousand samples, with even fewer containing a substantial number of anomalies. For this purpose, we generate $100$ synthetic datasets, each consisting of ten-dimensional data where each feature is normalized to range from $0$ to $1$. 

\begin{figure}
\centering
\includegraphics[width=0.799\textwidth]{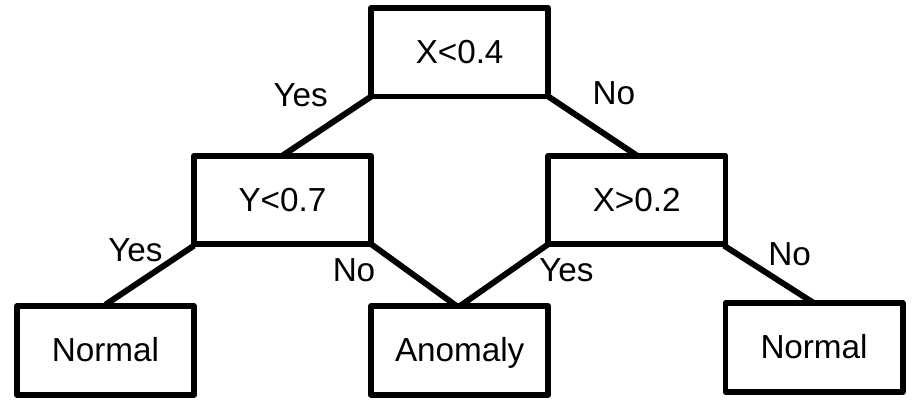}
\caption{Example decision tree used to generate our data. This tree is two dimensional and has a depth of $3$.}
\label{fig:tree}
\end{figure}

\begin{figure}
\centering
\includegraphics[width=0.799\textwidth]{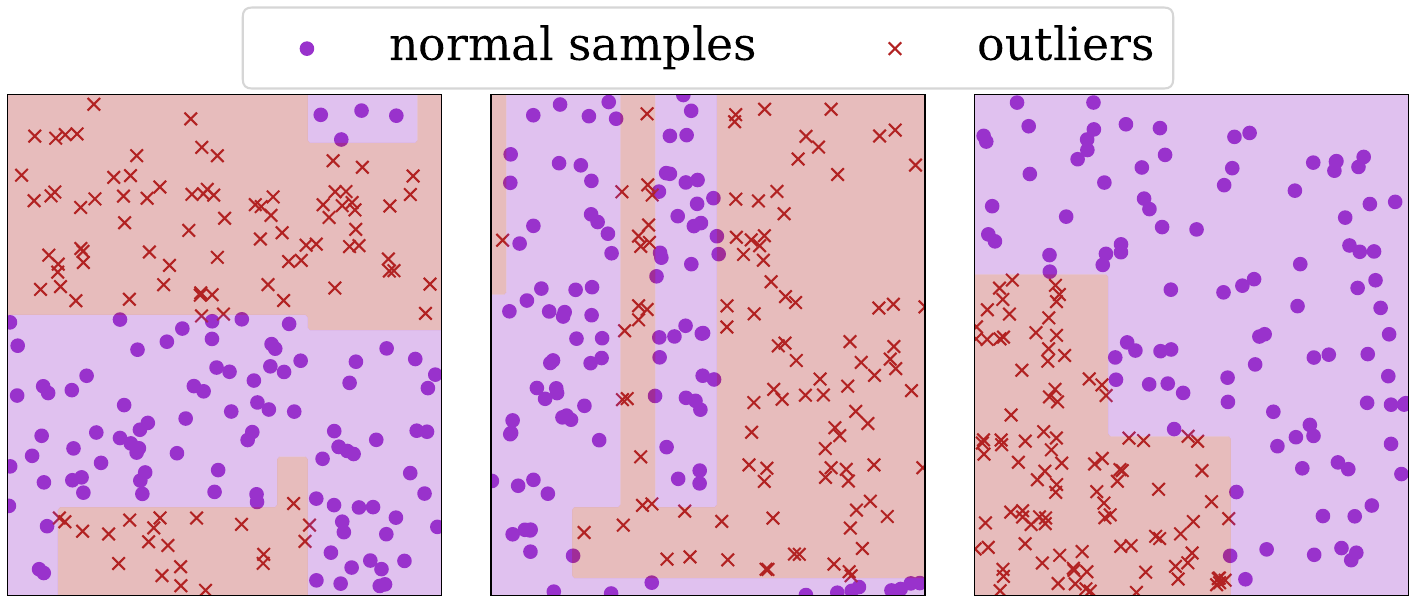}
\caption{Three examples of datasets generated using our data generation algorithm, shown in two dimensions. This setup enables us to create arbitrarily large datasets for our experiments. For this plot, we generate unusually many anomalies, to better visualize their distribution.}
\label{fig:example}
\end{figure}

Data generation proceeds by constructing a complete binary decision tree of depth $5$. In each branch of the tree, a feature and a split value are randomly selected, and each leaf node is labeled as either True or False. A sample is labeled as "normal" if it reaches a leaf marked True and "anomalous" otherwise, resulting in a complex yet randomized distribution of normal samples within a defined region. An example of such a decision tree is shown in Figure~\ref{fig:tree}. Also Figure~\ref{fig:example} illustrates three examples of these generated datasets (shown in two dimensions for visualization), demonstrating the variability and complexity of our data generation process.

For each dataset, we generate up to one million samples for the training set, as well as for both the normal and anomalous portions of the test set.

Using these synthetic datasets, we train various anomaly detection algorithms to detect significant differences between clean training data and contaminated test data. Each model is trained on $N$ normal samples and subsequently evaluated on a test set of $N$ samples, of which $\nu \cdot N$ samples are anomalous (with $\nu$ as the contamination rate). 

After training, we measure the effectiveness of anomaly detection by evaluating the fraction of datasets where a significant difference between normal and contaminated data is detected by a statistical test (with $p < 0.05$). We repeat this process across different anomaly detection algorithms, statistical tests, dataset sizes $N$, and contamination rates $\nu$ to comprehensively assess performance.


\section{Experiments}

\begin{figure}
\centering
\includegraphics[width=0.799\textwidth]{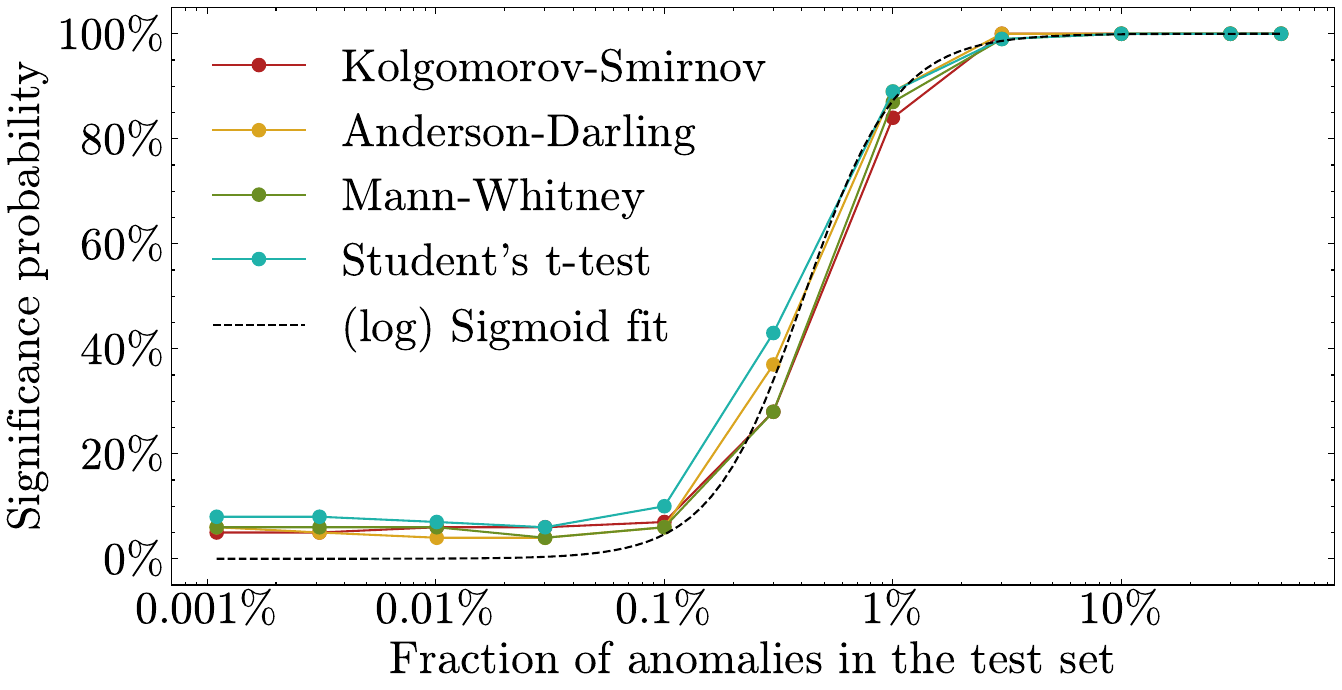}
\caption{Fraction of datasets that contain a significant difference between uncontaminated training data and contaminated test data for various statistical tests and contamination rates using a Gaussian Mixture Model with one million total points in the training and test sets.}
\label{fig:sigmoid}
\end{figure}

We begin by analyzing the fraction of datasets with significant differences on the maximum dataset size $N=1,000,000$ using a Gaussian Mixture Model (GMM~\cite{gmm}) in Figure~\ref{fig:sigmoid}.

Two key observations arise here. First, the difference between the applied statistical tests is minimal, with the largest difference observed between the Student's t-test and the Mann-Whitney test, approximately $15\%$ at $\nu=3\cdot 10^{-3}$. This holds across other values of $N$ and different anomaly detection algorithms, with a maximum observed deviation of $22\%$ between two statistical tests in all experiments studied in this paper. Therefore, for simplicity, we use the average significance probability over all tests in subsequent analyses. Second, the general curve appears to follow a sigmoid function in logarithmic space. Specifically, we fit a logarithmic sigmoid function to the average probability:

\begin{equation}
    P = \text{sigmoid}\left(\text{slope} \cdot (\log(x) - \log(\text{center}))\right)
    \label{eqn:logmoid}
\end{equation}

Using gradient descent, we estimate $\text{slope} \approx 2.15$ and $\text{center} \approx 4.1 \cdot 10^{-3}$ in this example relation. This sigmoidal function seems to capture the relationship well, suggesting that only two cases exist: at high contamination rates $\nu$, the difference between training and test data becomes statistically noticeable, while at low $\nu$, it is statistically negligible. The transition between these states is relatively sudden, spanning roughly a factor of $10$ in $\nu$. This pattern is consistent across different algorithms and values of $N$. Notably, however, we will show in the remainder of this paper, that the threshold for transition ($center$) depends on $N$ and the algorithm used.

\begin{figure}
\centering
\includegraphics[width=0.799\textwidth]{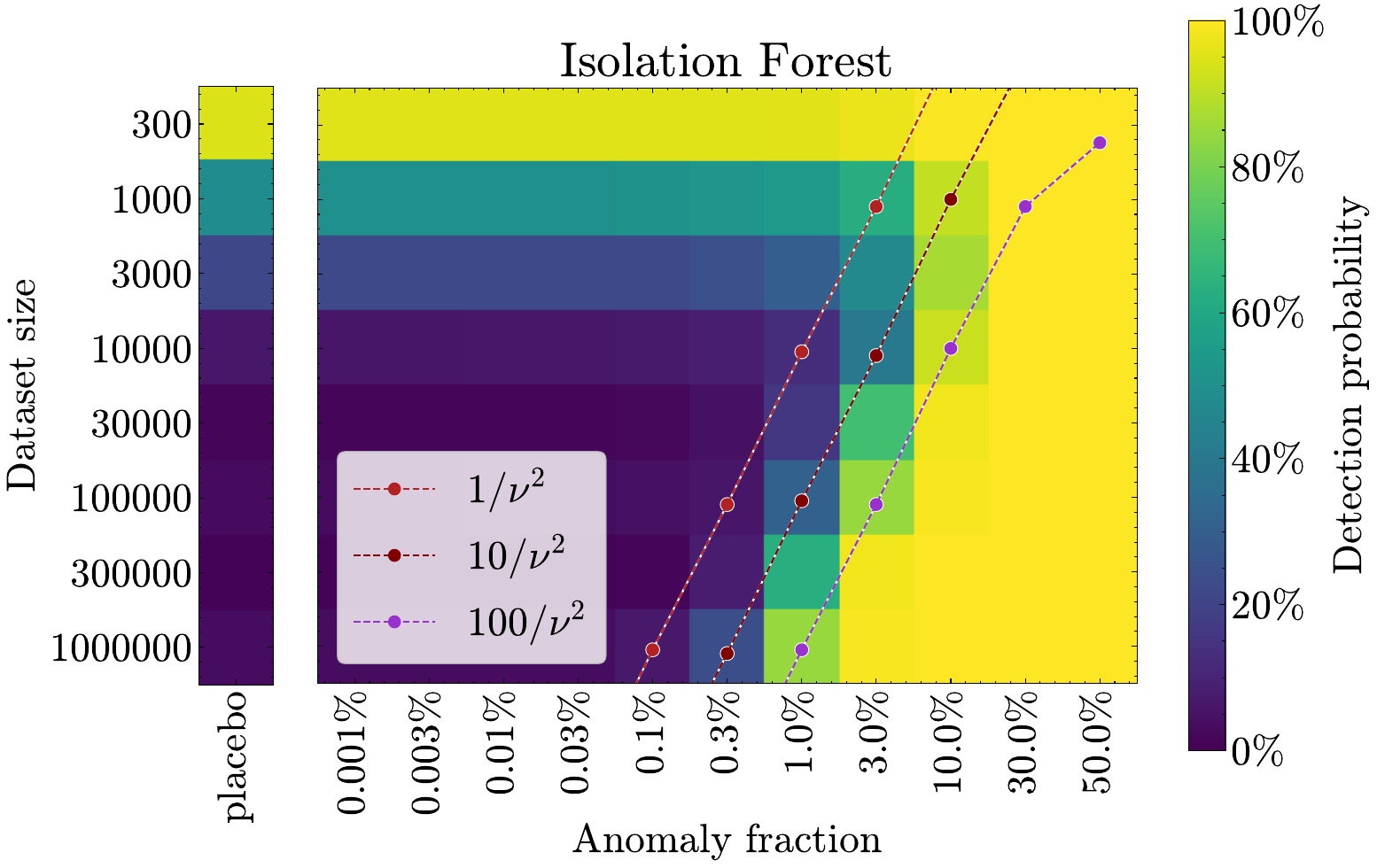}
\caption{Average detection probability using an Isolation Forest as a function of contamination fraction $\nu$ and dataset size $N$. "Placebo" represents the limit of no contamination ($\nu=0$).}
\label{fig:trice_ifor}
\end{figure}

\begin{figure}[h]
    \centering
    \begin{subfigure}[b]{0.49\textwidth}
        \includegraphics[width=\textwidth]{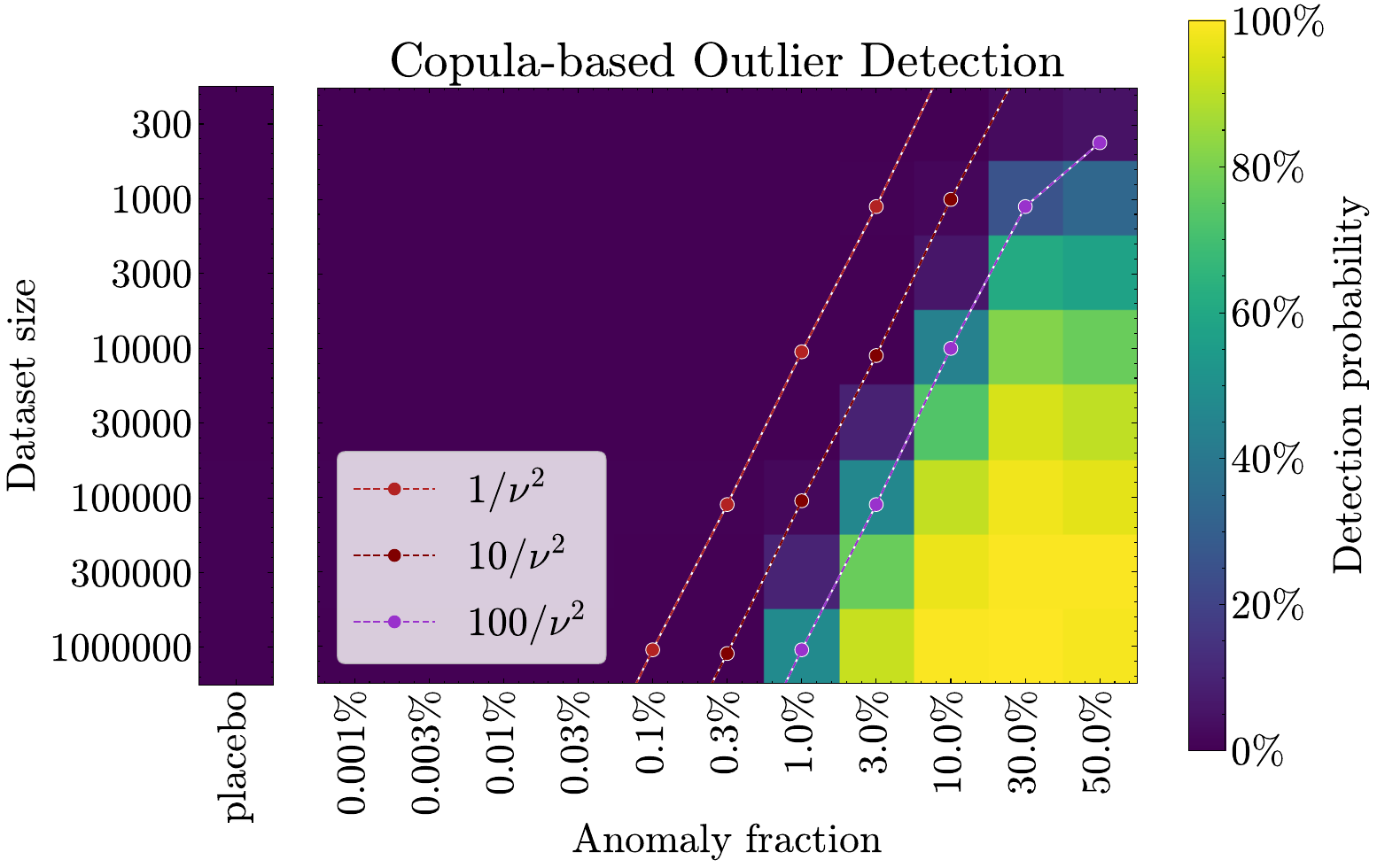}
        \caption{COPOD}
        \label{fig:copod}
    \end{subfigure}
    \begin{subfigure}[b]{0.49\textwidth}
        \includegraphics[width=\textwidth]{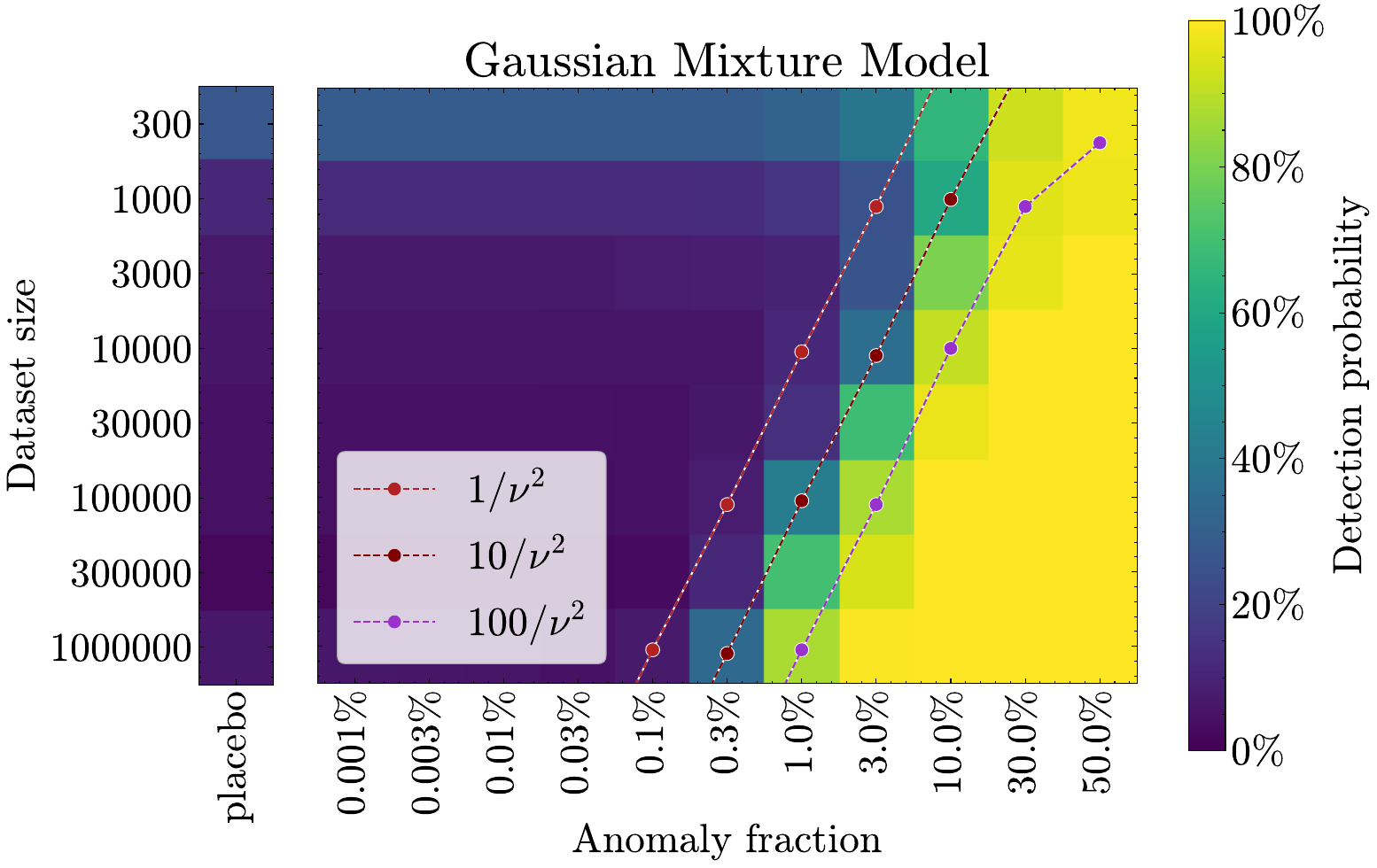}
        \caption{GMM}
        \label{fig:gmm}
    \end{subfigure}
    \\
    \begin{subfigure}[b]{0.49\textwidth}
        \includegraphics[width=\textwidth]{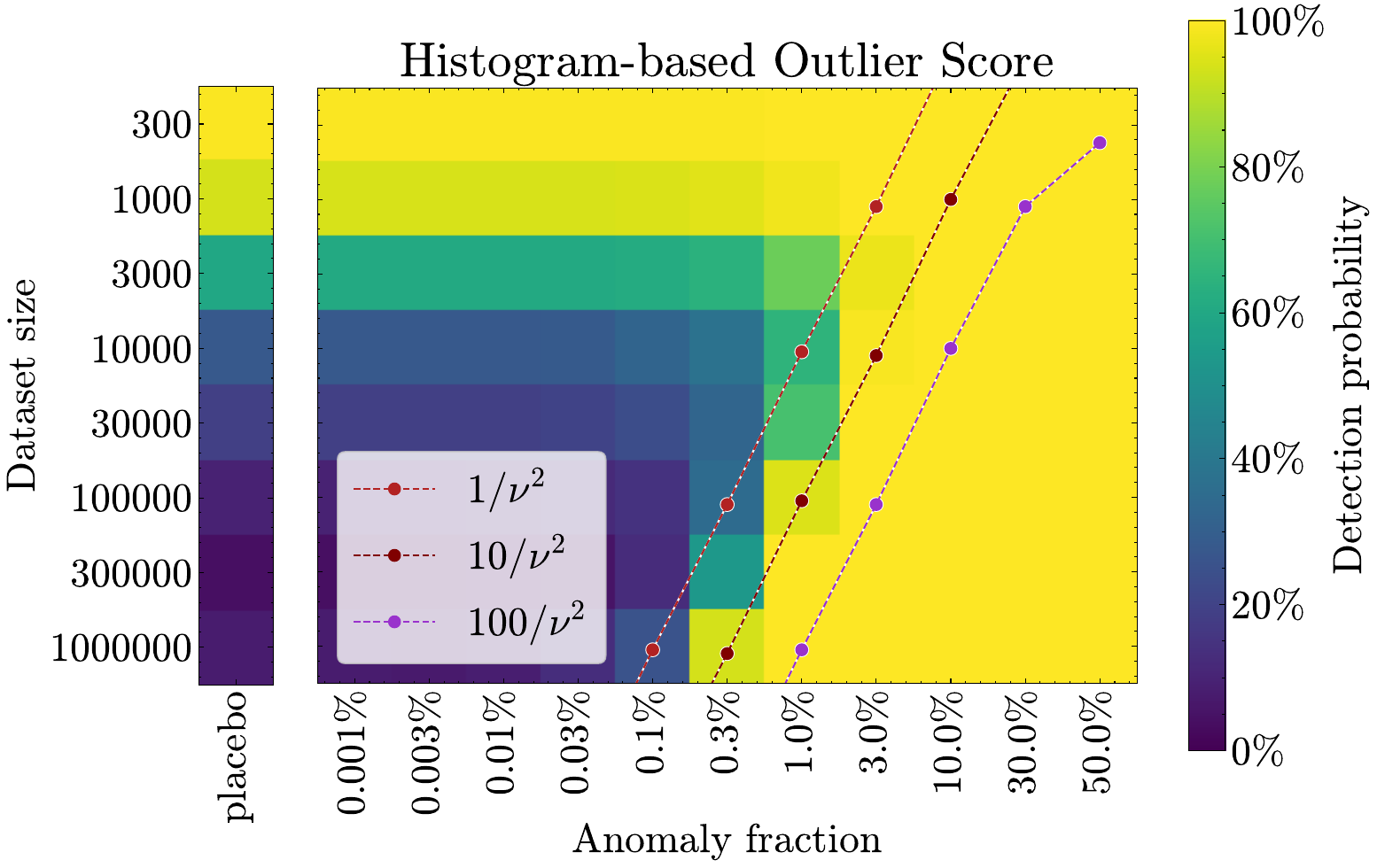}
        \caption{HBOS}
        \label{fig:hbos}
    \end{subfigure}
    \begin{subfigure}[b]{0.49\textwidth}
        \includegraphics[width=\textwidth]{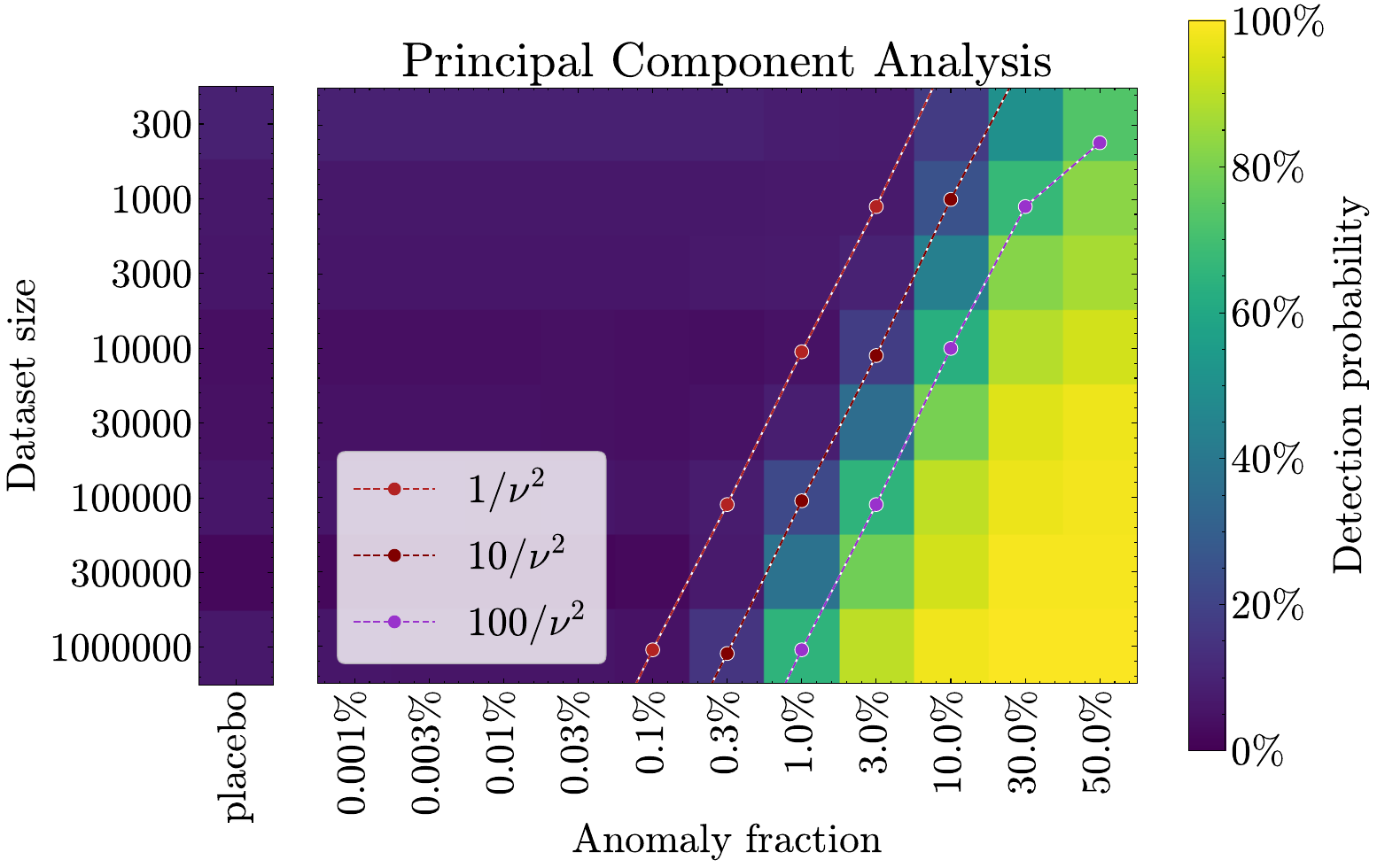}
        \caption{PCA}
        \label{fig:pca}
    \end{subfigure}
    \caption{Average detection probability as a function of contamination fraction $\nu$ and dataset size $N$ for various algorithms used in this study.}
    \label{fig:square_grid}
    
\end{figure}

To explore this, we plot the average detection probability as a function of both $N$ and $\nu$ for the Isolation Forest (IFor~\cite{ifor}) algorithm in Figure~\ref{fig:trice_ifor}. We include a "placebo" column, which indicates the likelihood of a false positive at $\nu=0$. Similar plots for other algorithms used in this study are shown in Figure~\ref{fig:square_grid}.

\begin{figure}
\centering
\includegraphics[width=0.799\textwidth]{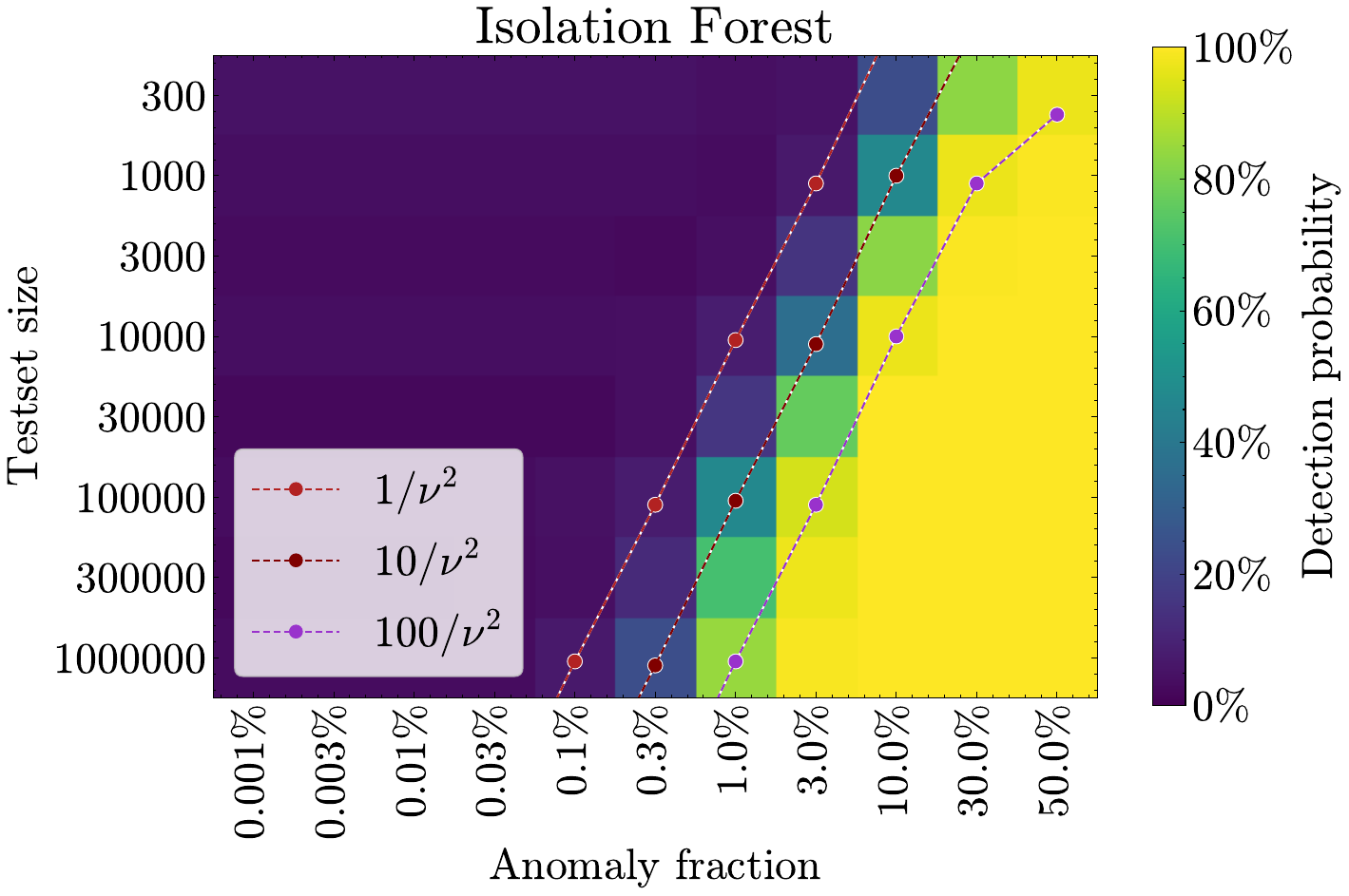}
\caption{Analogous to Figure~\ref{fig:trice_ifor} with maximum training set size but varying test set sizes.}
\label{fig:os}
\end{figure}

These plots reveal two main insights. First, for small dataset sizes, the likelihood of false positives (positive statistical tests in the placebo column) is substantial. This implies that small datasets cannot reliably detect distributional differences, as inherent variations exist even between samples from the same distribution. To confirm this, we conducted a complementary experiment with a fixed training set size of one million samples but varying test set sizes (Figure~\ref{fig:os}). As expected, false positives diminish in the larger test sets.

\begin{figure}
\centering
\includegraphics[width=0.799\textwidth]{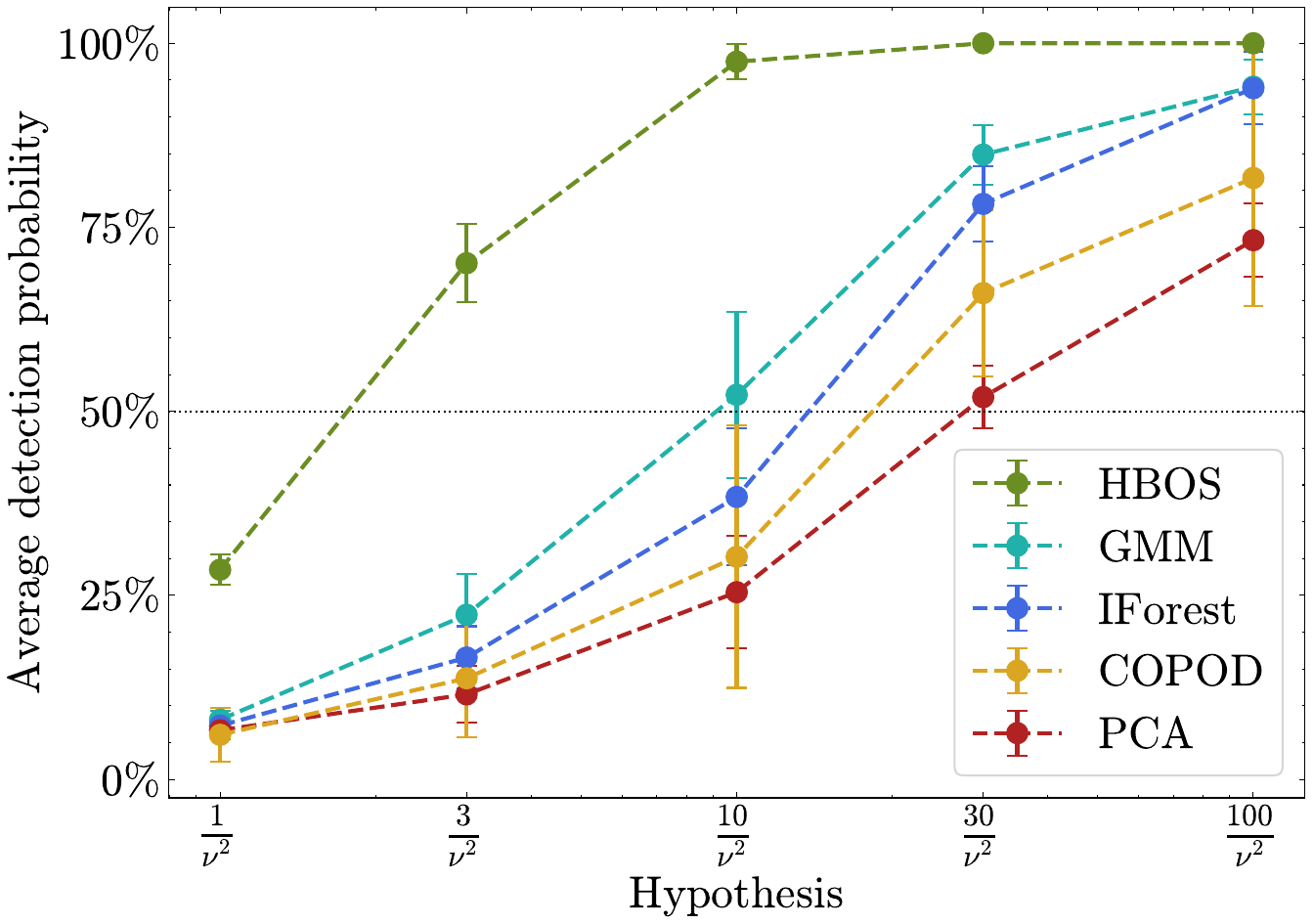}
\caption{Average detection probability for different algorithms and decision boundary hypotheses. Each value represents the average z/color value along the lines shown in Figure~\ref{fig:os}.}
\label{fig:running}
\end{figure}

The second insight is that the minimum anomaly fraction required to detect a training-test difference (i.e., $center$ in Equation~\ref{eqn:logmoid}) depends on dataset size. Specifically, the $50\%$ decision threshold follows a quadratic relation:

\begin{equation}
    N \ge \frac{\alpha_{\text{algo}}}{\nu^2}
    \label{eqn:hypo}
\end{equation}

where $\alpha_{\text{algo}}$ is an algorithm-specific constant. To further explore this dependency, we examine the average detection probability along these proposed decision boundaries for different algorithms and values of $\alpha_{\text{algo}}$ in Figure~\ref{fig:running}. We use here a constant training size to minimize the effects of false positives.

Similar to Figure~\ref{fig:sigmoid}, the detection probability curves generally also fit the logarithmic sigmoid form (Equation~\ref{eqn:logmoid}) with low uncertainties along the hypotheses lines, validating Equation~\ref{eqn:hypo}. Some algorithms exhibit higher detection probabilities (lower $\alpha_{\text{algo}}$), with the best (HBOS, $\alpha_{HBOS} \approx 2$) aligning well with our data's linear decision boundaries (Figure~\ref{fig:example}), while the worst (PCA, $\alpha_{PCA} \approx 30$) struggles to find a (often non-existent) lower dimensional representation. That different algorithms result in different $\alpha_{\text{algo}}$, encoding their ability to detect anomalies well constants suggests that on large, realistic datasets, such significance studies could serve as benchmarks.
Notably,  compared to the usual anomaly detection metrics, this approach requires no anomaly labels, enabling the benchmarking of algorithms on unlabeled (often significantly larger) datasets. Still, such an analysis might be quite expensive, as we also had to restrict our analysis to only very fast algorithms here.

Lastly, $\alpha_{\text{algo}}$ varies by an order of magnitude, which is relatively stable compared to the five orders of magnitude covered by $\nu$ values. Using that $\alpha$ has a limited range, we can estimate the minimum samples needed for $50\%$ detection probability, as shown in Table~\ref{tab:values}.

\begin{table}[h!]
\centering
\begin{tabular}{|c|c|c|c|c|c|c|}
\hline
\( \nu \) & $50\%$ & $10\%$ & $5\%$ & $1\%$ & $0.5\%$ & $0.1\%$ \\
\hline
\( N (\alpha=1) \) & 4 & 100 & 400 & 10{,}000 & 40{,}000 & 1{,}000{,}000 \\
\( N (\alpha=10) \) & 40 & 1{,}000 & 4{,}000 & 100{,}000 & 400{,}000 & 10{,}000{,}000 \\
\( N (\alpha=100) \) & 400 & 10{,}000 & 40{,}000 & 1{,}000{,}000 & 4{,}000{,}000 & 100{,}000{,}000 \\
\hline
\end{tabular}
\caption{Minimum test samples required for $50\%$ anomaly detection probability across various contamination rates $\nu$ and algorithmic quality measures $\alpha$.}
\label{tab:values}
\end{table}

These sample requirements are high, especially for low values of $\nu$. This is a fundamental challenge, as anomalies are typically rare. For example, with $\nu=1\%$ and an optimistic guess of $\alpha=1$, at least $N \ge 10{,}000$ samples are needed for $50\%$ detection probability. This limits benchmarking utility in anomaly detection. A recent survey~\cite{surveyzhao} collected a set of $47$ classical benchmark datasets, of which only $14$ ($30\%$) contain over $10{,}000$ samples, required for successful detection with $\nu=1\%$ and $\alpha=1$. Most of these datasets contain a much higher fraction of anomalies ($\nu\gg1\%$), as this is beneficial for benchmarking and usually does not affect the ROC-AUC anomaly ranking. However, a contamination rate of $\nu>10\%$ is not realistic in anomaly detection applications. And even with this unrealistically high fraction of anomalies, only $55\%$ of datasets fulfill Equation~\ref{eqn:hypo} for $\alpha=10$.

\subsection{Thought Experiment}

We propose a thought experiment to provide intuition on the relationship between contamination rate and the minimum number of samples needed. For this, we consider a simplified version of our experiment, where anomalies are identifiable if their maximum anomaly score exceeds that of the normal samples. We assume that anomaly scores are generated from two Gaussian distributions, \( G_N \) and \( G_A \), with parameters \(\mu_A - \mu_N = 2\) and \(\sigma_A = \sigma_N = 1\). While scores from the abnormal distribution \( G_A \) are generally higher, our test may still fail if the number of normal samples (\( N_{normal} \)) significantly exceeds the number of anomalies (\( N_{abnormal} \)). Thus, a relationship exists between the count of anomalies and the maximum allowable normal samples. Here, we search for this relationship, denoted as \( N_{normal} \le f(N_{abnormal}) \).

\begin{figure}
\centering
\includegraphics[width=0.799\textwidth]{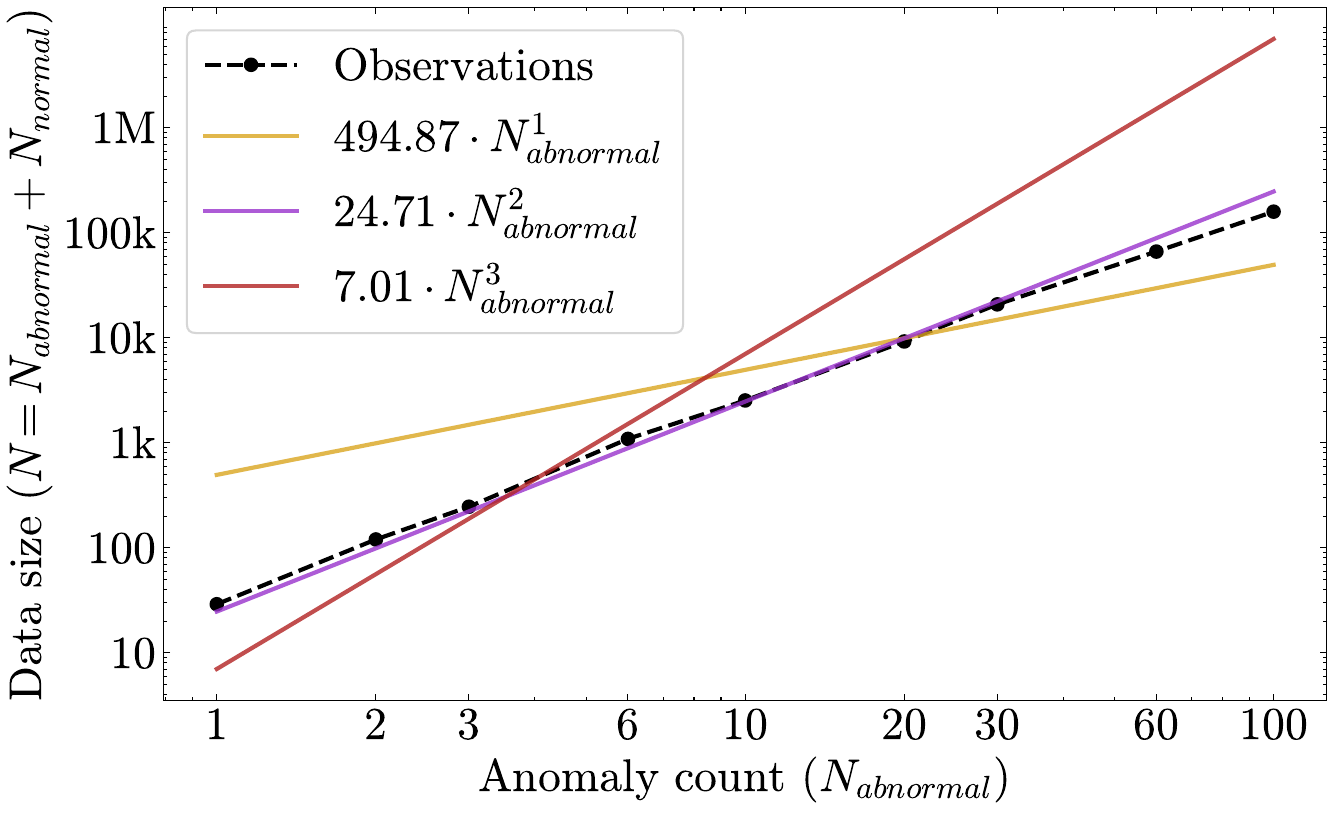}
\caption{Size of the dataset required to solve our thought experiment with a $50\%$ probability. Solving this involves a probabilistic optimization problem, leading to minor inaccuracies. Nonetheless, our measurement appears to closely follow a quadratic relationship.}
\label{fig:thought}
\end{figure}

To achieve this, we draw multiple sets of normal and abnormal samples for given values of \( N_{abnormal} \) and \( N_{normal} \), identifying the value of \( N_{normal}(N_{abnormal}) \) where exactly half of the sets ($\frac{3000}{2}=1500$) succeed. This is accomplished using a logarithmic binary search. Given the random nature of these variables, this optimization is somewhat imprecise; therefore, we employ a large sample size ($3000$ distributions) at each optimization step and repeat the optimization to confirm results, finding only negligible differences. We repeat this search for various values of \( N_{abnormal} \), and illustrate the results in Figure~\ref{fig:thought}.

From this experiment, the quadratic monomial relationship appears to fit best:

\begin{equation}
    N_{\text{abnormal}} + N_{\text{normal}} \le \frac{1}{\alpha} \cdot N_{\text{abnormal}}^2
\end{equation}

This relationship aligns with our proposed Equation~\ref{eqn:hypo}, suggesting that our proposed relation is broadly applicable and extends beyond the use of statistical tests.

\begin{proof}
Using \(\nu = \frac{N_{\text{abnormal}}}{N_{\text{normal}} + N_{\text{abnormal}}}\), we can rewrite Equation~\ref{eqn:hypo} as follows:

\[
N = N_{\text{abnormal}} + N_{\text{normal}} \ge \alpha \cdot \frac{(N_{\text{abnormal}} + N_{\text{normal}})^2}{N_{\text{abnormal}}^2}
\]

\[
\Rightarrow N_{\text{abnormal}}^2 \ge \alpha \cdot (N_{\text{abnormal}} + N_{\text{normal}}) \;\Leftrightarrow\; N_{\text{abnormal}} + N_{\text{normal}} \le \frac{1}{\alpha} \cdot N_{\text{abnormal}}^2
\]


$\qed$
\end{proof}

\section{Conclusion}

In this paper, we present the first explicit study addressing the fundamental question of whether any anomalies exist within a dataset. Through over three million statistical tests across various anomaly detection tasks, algorithms, and statistical methods, we establish a clear requirement for reliably detecting the existence of anomalies. Given the size of an unlabeled dataset \( N \), the contamination rate \( \nu \), and an algorithm-dependent constant \( \alpha_{\text{algo}} \), we find that

\[
    N = N_{normal}+N_{abnormal}\ge \frac{\alpha_{\text{algo}}}{\nu^2} \hspace{2em}\Leftrightarrow\hspace{2em} N_{\text{normal}} \le \frac{1}{\alpha_{\text{algo}}} \cdot N_{\text{abnormal}}^2
\]

This condition serves as an upper bound on anomaly rarity, beyond which it becomes infeasible to prove the existence of anomalies.

For future work, we plan to explore the generality of this relationship further. For instance, we aim to conduct user studies to examine whether human cognition aligns with this mathematical law. Depending on the findings, this limitation may be alleviated by advancing weakly supervised anomaly detection, potentially utilizing large, unrelated datasets (self-supervised anomaly detection~\cite{selfsupervised}).

%
%
%
\bibliographystyle{splncs04}
\bibliography{one,new}
\end{document}